\documentclass{article}
\usepackage{graphicx}

\newcommand{\eq}[1]{Eq.~(\ref{eq.#1})} 

\newcommand{\fig}[1]{Fig.~\ref{fig.#1}}

\newcommand{\tbl}[1]{Table~\ref{table.#1}}

\newcommand{\sect}[1]{Section \ref{sect.#1}}

\newcommand{\sectlabel}[1]{\label{sect.#1}}
\newcommand{\eqlabel}[1]{\label{eq.#1}}
\newcommand{\figlabel}[1]{\label{fig.#1}}
\newcommand{\tbllabel}[1]{\label{table.#1}}

\newcommand{\figwidth}{4in}
\newcommand{\Poisson}{\mbox{Po}}
\newcommand{\Reynolds}{\mbox{Re}}
\newcommand{\vAvg}{v_{\rm{avg}}}

\newcommand{\Ftarget}{F_{\rm{source}}}
\newcommand{\Ltarget}{L_{\rm{source}}}
\newcommand{\DRobot}{D_{\rm{robot}}}
\newcommand{\rhoRobot}{\rho_{\rm{robot}}} 
\newcommand{\rhoVessel}{\rho_{\rm{vessel}}} 
\newcommand{\Vtissue}{V} 
\newcommand{\vFluid}{{\bf v}}
\newcommand{\position}{{\bf r}}
\newcommand{\Flux}{{\bf F}}
\newcommand{\sensorRate}{\Omega_{\rm{robot}}} 
\newcommand{\sensorRateOne}{\omega_{\rm{robot}}} 
\newcommand{\sourceRate}{\omega_{\rm{source}}} 
\newcommand{\backgroundRate}{\omega_{\rm{background}}} 
\newcommand{\Ebackground}{\mu_{\rm{background}}} 

\newcommand{\Csource}{C_{\rm{source}}}
\newcommand{\Cbackground}{C_{\rm{background}}}

\newcommand{\BoltzmannConstant}{\ensuremath{k_B}}

\newcommand{\Tmeasure}{T_{\rm{measure}}}
\newcommand{\Cthreshold}{C_{\rm{threshold}}}
\newcommand{\Ttask}{T} 

\newcommand{\Rmonitor}{R_{\rm{m}}} 

\newcommand{\aTransition}[2]{{\alpha_{\rm{#1} \rightarrow \rm{#2}}}}
\newcommand{\aMonitorToDetected}{\aTransition{m}{d}}

\newcommand{\meter}{\mbox{m}}
\newcommand{\millimeter}{\mbox{mm}}
\newcommand{\micron}{\mbox{$\mu$m}}
\newcommand{\second}{\mbox{s}}
\newcommand{\millisecond}{\mbox{ms}}

\newcommand{\molecule}{\mbox{molecule}}

\newcommand{\robot}{\mbox{robot}}

\newcommand{\picoWatt}{\mbox{pW}}
\newcommand{\Joule}{\mbox{J}}

\begin{document}
\title{Distributed Control of Microscopic Robots in Biomedical Applications}
\author{Tad Hogg\\Hewlett-Packard Laboratories\\Palo Alto, CA}

\maketitle

\begin{abstract}
Current developments in molecular electronics, motors and chemical
sensors could enable constructing large numbers of devices able to
sense, compute and act in micron-scale environments. Such
microscopic machines, of sizes comparable to bacteria, could
simultaneously monitor entire populations of cells individually in
vivo. This paper reviews plausible capabilities for microscopic
robots and the physical constraints due to operation in fluids at
low Reynolds number, diffusion-limited sensing and thermal noise
from Brownian motion. Simple distributed controls are then
presented in the context of prototypical biomedical tasks, which
require control decisions on millisecond time scales. The
resulting behaviors illustrate trade-offs among speed, accuracy
and resource use. A specific example is monitoring for patterns of
chemicals in a flowing fluid released at chemically distinctive
sites. Information collected from a large number of such devices
allows estimating properties of cell-sized chemical sources in a
macroscopic volume. The microscopic devices moving with the fluid
flow in small blood vessels can detect chemicals released by
tissues in response to localized injury or infection. We find the
devices can readily discriminate a single cell-sized chemical
source from the background chemical concentration, providing
high-resolution sensing in both time and space. By contrast, such
a source would be difficult to distinguish from background when
diluted throughout the blood volume as obtained with a blood
sample.

\end{abstract}

\newpage
\section{Microscopic Robots}

Robots with sizes comparable to bacteria could operate in
microscopic environments on the scale of individual cells in
biological organisms. Such robots are small enough to move through
the tiniest blood vessels, so could pass within a few cell
diameters of most cells in large organisms via their circulatory
systems to perform a wide variety of biological research and
medical tasks. For instance, robots and nanoscale-structured
materials inside the body could significantly improve disease
diagnosis and treatment~\cite{freitas99,morris01,nih03,keszler01}.
Initial tasks for microscopic robots include in vitro research via
simultaneous monitoring of chemical signals exchanged among many
bacteria in a biofilm. The devices could also operate in
multicellular organisms as passively circulating sensors. Such
devices, with no need for locomotion, would detect programmed
patterns of chemicals as they pass near cells. More advanced
technology could create devices able to communicate to external
detectors, allowing real-time in vivo monitoring of many cells.
The devices could also have capabilities to act on their
environment, e.g., releasing drugs at locations with specific
chemical patterns or mechanically manipulating objects for
microsurgery. Extensive development and testing is necessary
before clinical use, first for high-resolution diagnostics and
later for programmed actions at cellular scales.

Realizing these benefits requires fabricating the robots cheaply,
in large numbers and with sufficient capabilities. Such
fabrication is beyond current technology. Nevertheless, ongoing
progress in engineering nanoscale devices could eventually enable
production of such robots. One approach is engineering biological
systems, e.g., bacteria executing simple programs~\cite{weiss06},
and DNA computers responding to logical combinations of
chemicals~\cite{benenson04}. However, biological organisms have
limited material properties and computational speed. Instead we
focus on machines based on plausible extensions of current
molecular-scale electronics, sensors and
motors~\cite{berna05,collier99,craighead00,howard97,fritz00,montemagno99,soong00,wang05}.
These devices could provide components for stronger and faster
microscopic robots than is possible with biological organisms.

Because we cannot yet fabricate microscopic robots with molecular
electronics components, estimates of their performance rely on
plausible extrapolations from current technology. The focus in
this paper is on biomedical applications requiring only modest
hardware capabilities, which will be easier to fabricate than more
capable robots. Designing controls for microscopic robots is a key
challenge: not only enabling useful performance but also
compensating for their limited computation, locomotion or
communication abilities. Theoretical studies allow developing such
controls and estimating their performance prior to fabrication,
thereby indicating design tradeoffs among hardware capabilities,
control methods and task performance. Such studies of microscopic
robots complement analyses of individual nanoscale
devices~\cite{mccurdy02,wang05}, and indicate even modest
capabilities enable a range of novel applications.

The operation of microscopic robots differs significantly from
larger robots~\cite{mataric92}, especially for biomedical
applications. First, the physical environment is dominated by
viscous fluid flow. Second, thermal noise is a significant source
of sensor error and Brownian motion limits the ability to follow
precisely specified paths. Third, relevant objects are often
recognizable via chemical signatures rather than, say, visual
markings or specific shapes. Fourth, the tasks involve large
numbers of robots, each with limited abilities. Moreover, a task
will generally only require a modest fraction of the robots to
respond appropriately, not for all, or even most, robots to do so.
Thus controls using random variations are likely to be effective
simply due to the large number of robots. This observation
contrasts with teams of larger robots with relatively few members:
incorrect behavior by even a single robot can significantly
decrease team performance. These features suggest reactive
distributed control is particularly well-suited for microscopic
robots.

Organisms contain many microenvironments, with distinct physical,
chemical and biological properties. Often, precise quantitative
values of properties relevant for robot control will not be known
a priori. This observation suggests a multi-stage protocol for
using the robots. First, an information-gathering stage with
passive robots placed into the organism, e.g., through the
circulatory system, to measure relevant properties~\cite{hogg06b}.
The information from these robots, in conjunction with
conventional diagnostics at larger scales, could then determine
appropriate controls for further actions in subsequent stages of
operation.

For information gathering, each robot notes in its memory whenever
chemicals matching a prespecified pattern are found. Eventually,
the devices are retrieved and information in their memories
extracted for further analysis in a conventional computer with far
more computational resources than available to any individual
microscopic robot. This computer would have access to information
from many robots, allowing evaluation of aggregate properties of
the population of cells that individual robots would not have
access to, e.g., the number of cells presenting a specific
combination of chemicals. This information allows estimating
spatial structure and strength of the chemical sources. The robots
could detect localized high concentrations that are too low to
distinguish from background concentrations when diluted in the
whole blood volume as obtained with a sample. Moreover, if the
detection consists of the joint expression of several chemicals,
each of which also occurs from separate sources, the robot's
pattern recognition capability could identify the spatial
locality, which would not be apparent when the chemicals are mixed
throughout the blood volume.

Estimating the structure of the chemical sources from the
microscopic sensor data is analogous to computerized
tomography~\cite{natterer01}. In tomography, the data consists of
integrals of the quantity of interest (e.g., density) over a large
set of lines with known geometry selected by the experimenter. The
microscopic sensors, on the other hand, record data points
throughout the tissue, providing more information than just one
aggregate value such as the total number of events. However, the
precise path of each sensor through the tissue, i.e., which vessel
branches it took and the locations of those vessels, will not be
known. This mode of operation also contrasts with uses of larger
distributed sensor networks able to process information and
communicate results while in use.

Actions based on the information from the robots would form a
second stage of activity, perhaps with specialized microscopic
robots (e.g., containing drugs to deliver near cells), with
controls set based on the calibration information retrieved
earlier. For example, the robots could release drugs at chemically
distinctive sites~\cite{freitas99,freitas06} with specific
detection thresholds determined with the information retrieved
from the first stage of operation. Or robots could aggregate at
the chemical sources~\cite{casal03,hogg06a} or manipulate
biological structures based on surface chemical patterns on cells,
e.g., as an aid for microsurgery in repairing injured
nerves~\cite{hogg05}. These active scenarios require more advanced
robot capabilities, such as locomotion and communication, than
needed for passive sensing. The robots could monitor environmental
changes due to their actions, thereby documenting the progress of
the treatment. Thus the researcher or physician could monitor the
robots' progress and decide whether and when they should continue
to the next step of the procedure. Using a series of steps, with
robots continuing with the next step only when instructed by the
supervising person, maintains overall control of the robots, and
simplifies the control computations each robot must perform
itself.

To illustrate controls for large collections of microscopic
robots, this paper considers a prototypical diagnostic task of
finding a small chemical source in a multicellular organism via
the circulatory system. To do so, we first describe plausible
capabilities for the robots and techniques for evaluating their
behavior. We then examine a specific task scenario. The emphasis
here is on feasible performance with plausible biophysical
parameters and robot capabilities. Evaluation metrics include
minimizing hardware capabilities to simplify fabrication and
ensuring safety, speed and accuracy for biological research or
treatment in a clinical setting.

\section{Capabilities of Microscopic Robots}

This section describes plausible robot capabilities based on
currently demonstrated nanoscale technology. Minimal capabilities
needed for biomedical tasks include chemical sensing, computation
and power. Additional capabilities, enabling more sophisticated
applications, include communication and locomotion.

\subsection{Chemical Sensing}

Large-scale robots often use sonar or cameras to sense their
environment. These sensors locate objects from a distance, and
involve sophisticated algorithms with extensive computational
requirements. In contrast, microscopic robots for biological
applications will mainly use chemical sensors, e.g., the selective
binding of molecules to receptors altering the electrical
characteristics of nanoscale wires. The robots could also examine
chemicals inside nearby cells~\cite{sunney06}.

Microscopic robots and bacteria face similar physical constraints
in detecting chemicals~\cite{berg77}. The diffusive capture rate
$\gamma$ for a sphere of radius $a$ in a region with concentration
$C$ is~\cite{berg93}
\begin{equation}\eqlabel{diffusive capture}
\gamma = 4 \pi D a C
\end{equation}
where $D$ is the diffusion coefficient of the chemical. Even when
sensors cover only a relatively small fraction of the device
surface, the capture rate is almost this large~\cite{berg93}.
Nonspherical devices have similar capture rates so \eq{diffusive
capture} is a reasonable approximation for a variety of designs.

Current molecular electronics~\cite{wang05} and nanoscale
sensors~\cite{patolsky05,sheehan05} indicate plausible sensor
capabilities. At low concentrations, sensor performance is
primarily limited by the time for molecules to diffuse to the
sensor and statistical fluctuations in the number of molecules
encountered is a major source of noise.

\subsection{Timing and Computation}

With the relevant fluid speeds and chemical concentrations
described in \sect{task}, robots pass through high concentrations
near individual cells on millisecond time scales. Thus identifying
significant clusters of detections due to high concentrations
requires a clock with millisecond resolution. This clock need not
be globally synchronized with other devices.

In a simple scenario, devices just store sensor detections in
their memories for later retrieval. In this case most of the
computation to interpret sensor observations takes place in larger
computers after the devices are retrieved. Recognizing and storing
a chemical detection involves at least a few arithmetic operations
to compare sensor counts to threshold values stored in memory. An
estimate on the required computational capability is about 100
elementary logic operations and memory accesses within a
$10\millisecond$ measurement time. This gives about $10^4$ logic
operations per second. While modest compared to current computers,
this rate is significantly faster than demonstrated for
programmable bacteria~\cite{weiss06} but well within the
capabilities of molecular electronics.

\subsection{Communication}

A simple form of one-way communication is robots passively sensing
electromagnetic or acoustic signals from outside the body. Such
signals could activate robots only within certain areas of the
body at, say, centimeter length scales. Additional forms of
communication, between nearby devices and sending information to
detectors outside the organism, are more difficult to fabricate
and require significant additional power.

Devices with chemical sensors could communicate via chemical
signals. Such diffusion-mediated signals are not effective for
communicating over distances beyond a few microns but could mark
the environment for detection by other robots that pass nearby
later, i.e., stigmergy~\cite{bonabeau99}. Acoustic signals provide
more versatile communication. Compared to fluid flow, acoustic
signals are essentially instantaneous, but power constraints limit
their range to about $100\micron$~\cite{freitas99}.

\subsection{Locomotion}

Biomedical applications will often involve robots operating in
fluids. Viscosity dominates the robot motion, with different
physical behaviors than for larger organisms and
robots~\cite{purcell77,vogel94,fung97,karniadkis02,squires05}. The
ratio of inertial to viscous forces for an object of size $s$
moving with velocity $v$ through a fluid with viscosity $\eta$ and
density $\rho$ is the Reynolds number $\Reynolds \equiv s \rho v /
\eta$. Using typical values for density and viscosity (e.g., of
water or blood plasma) in \tbl{parameters} and noting that
reasonable speeds for robots with respect to the
fluid~\cite{freitas99} are comparable to the fluid flow speed in
small vessels, i.e., $\sim 1\mbox{mm/\second}$, motion of a
1-micron robot has $\Reynolds \approx 10^{-3}$, so viscous forces
dominate. Consequently, robots applying a locomotive force quickly
reach terminal velocity in the fluid, i.e., applied force is
proportional to velocity rather than the more familiar
proportionality to acceleration of Newton's law $F = m a$. By
contrast, a swimming person has $\Reynolds$ about a billion times
larger.

Flow in a pipe of uniform radius $R$ has a parabolic velocity
profile: velocity at distance $r$ from the axis is
\begin{equation}\eqlabel{fluid velocity}
v(r) = 2 \vAvg (1-(r/R)^2)
\end{equation}
where $\vAvg$ is the average speed of fluid in the pipe.

Robots moving through the fluid encounter significant drag. For
instance, an isolated sphere of radius $a$ moving at speed $v$
through a fluid with viscosity $\eta$ has a drag force
\begin{equation}\eqlabel{drag force}
6 \pi a \eta v
\end{equation}
Although not quantitatively accurate near boundaries or other
objects, this expression estimates the drag in those cases as
well. For instance, a numerical evaluation of drag force on a
$1\micron$-radius sphere moving at velocity $v$ with respect to
the fluid flow near the center of a $5\micron$-radius pipe, has
drag about three times larger than given by \eq{drag force}. Other
reasonable choices for robot shape have similar drag.

Fluid drag moves robots in the fluid. An approximation is robots
without active locomotion move with the same velocity as fluid
would have at the center of the robot if the robot were not there.
Numerical evaluation of the fluid forces on the robots for the
parameters of \tbl{parameters} show the robots indeed move close
to this speed when the spacing between robots is many times their
size. Closer packing leads to more complex motion due to
hydrodynamic interactions~\cite{hernandez05,riedel05}.

\subsection{Additional Sensing Capabilities}

In addition to chemical sensing, robots could sense other
properties to provide high-resolution spatial correlation of
various aspects of their environment. For example, nanoscale
sensors for fluid motion can measure fluid flow rates at speeds
relevant for biomedical tasks~\cite{ghosh03}, allowing robots to
examine in vivo microfluidic behavior in small vessels. In
particular, at low Reynolds number, boundary effects extend far
into the vessel~\cite{squires05}, giving an extended gradient in
fluid speed with higher fluid shear rates nearer the wall. Thus,
several such sensors, extending a small distance from the device
surface in various directions, could estimate shear rates and
hence the direction to the wall or changes in the vessel geometry.
Another example for additional sensing is optical scattering in
cells, as has been demonstrated to distinguish some cancer from
normal cells in vitro~\cite{gourley05}.

\subsection{Power}

To estimate the power for robot operation, each logic operation in
current electronic circuits uses $10^4 - 10^5$ times the thermal
noise level $\BoltzmannConstant T = 4 \times 10^{-21}\Joule$ at
the fluid temperature of \tbl{parameters}, where
$\BoltzmannConstant$ is the Boltzmann constant. Near term
molecular electronics could reduce this to $\approx 10^3
\BoltzmannConstant T$, in which case $10^4$ operations per second
uses a bit less than $0.1\picoWatt$.  Additional energy will be
needed for signals within the computer and with its memory.

This power is substantially below the power required for
locomotion or communication. For instance, due to fluid drag and
the inefficiencies of locomotion in viscous fluids, robots moving
through the fluid at $\approx 1\mbox{mm}/\second$ dissipate a
picowatt~\cite{berg93}. However, these actions may operate only
occasionally, and for short times, when the sensor detects a
signal, whereas computation could be used continuously while
monitoring for such signals.

For tasks of limited duration, an on-board fuel source created
during manufacture could suffice. Otherwise, the robots could use
energy available in their environment, such as converting
vibrations to electrical energy~\cite{wang06} or chemical
generators. Typical concentrations of glucose and oxygen in the
bloodstream could generate $\approx 1000\picoWatt$ continuously,
limited primarily by the diffusion rate of these molecules to the
device~\cite{freitas99}. For comparison, a typical person at rest
uses about $100$ watts.

\section{Evaluating Collective Robot Performance}

Because the microscopic robots can not yet be fabricated and
quantitative biophysical properties of many microenvironments are
not precisely known, performance studies must rely on plausible
models of both the machines and their task
environments~\cite{drexler92,freitas99,requicha03}.
Microorganisms, which face physical microenvironments similar to
those of future microscopic robots, give some guidelines for
feasible behaviors.

Cellular automata are one technique to evaluate collective robot
behavior. For example, a two-dimensional scenario shows how robots
could assemble structures~\cite{arbuckle04} using local rules.
However, cellular automata models either ignore or greatly
simplify physical behaviors such as fluid flow. Another analysis
technique considers swarms~\cite{bonabeau99}, which are
well-suited to microscopic robots with their limited physical and
computational capabilities and large numbers. Most swarm studies
focus on macroscopic robots or behaviors in abstract
spaces~\cite{gazi04} which do not specifically include physical
properties unique to microscopic robots. In spite of the
simplified physics, these studies show how local interactions
among robots lead to various collective behaviors and provide
broad design guidelines.

Simulations including physical properties of microscopic robots
and their environments can evaluate performance of robots with
various capabilities. Simple models, such as a two-dimensional
simulation of chemotaxis~\cite{dhariwal04}, provide insight into
robots find microscopic chemical sources. A more elaborate
simulator~\cite{cavalcanti02} includes three-dimensional motions
in viscous fluids, Brownian motion and environments with numerous
cell-sized objects, though without accounting for how they change
the fluid flow. Studies of hydrodynamic
interactions~\cite{hernandez05} among moving devices include more
accurate fluid effects.

Another approach to robot behaviors employs a stochastic
mathematical framework for distributed computational
systems~\cite{hogg04b,lerman01}. This method directly evaluates
average behaviors of many robots through differential equations
determined from the state transitions used in the robot control
programs. Direct evaluation of average behavior avoids the
numerous repeated runs of a simulation needed to obtain the same
result. This approach is best suited for simple control
strategies, with minimal dependencies on events in individual
robot histories. Microscopic robots, with limited computational
capabilities, will likely use relatively simple reactive controls
for which this analytic approach is ideally suited. Moreover,
these robots will often act in environments with spatially varying
fields, such as chemical concentrations and fluid velocities. Even
at micron scales, the molecular nature of these quantities can be
approximated as continuous fields with behavior governed by
partial differential equations. For application to microscopic
robots, this approximation extends to the robots themselves,
treating their locations as a continuous concentration field, and
their various control states as corresponding to different fields,
much as multiple reacting chemicals are described by separate
concentration fields. This continuum approximation for average
behavior of the robots will not be as accurate as when applied to
chemicals or fluids, but nevertheless gives a simple approach to
average behaviors for large numbers of robots responding to
spatial fields. One example of this approach is following chemical
gradients in one dimension without fluid flow~\cite{galstyan05}.

Cellular automata, swarms, physically-based simulations and
stochastic analysis are all useful tools for evaluating the
behaviors of microscopic robots. One example is evaluating the
feasibility of rapid, fine-scale response to chemical events too
small for detection with conventional methods, including sensor
noise inherent in the discrete molecular nature of low
concentrations. This paper examines this issue in a prototypical
task using the stochastic analysis approach. This method allows
incorporating more realistic physics than used with cellular
automata studies, and is computationally simpler than repeated
simulations to obtain average behaviors. This technique is limited
in requiring approximations for dependencies introduced by the
robot history, but readily incorporates physically realistic
models of sensor noise and consequent mistakes in the robot
control decisions. The stochastic analysis indicates plausible
performance, on average, and thereby suggests scenarios suited for
further, more detailed, simulation studies.

\section{A Task Scenario}\sectlabel{task}

As a prototypical task for microscopic robots, we consider their
ability to respond to a cell-sized source releasing chemicals into
a small blood vessel. This scenario illustrates a basic capability
for the robots: identifying small chemically-distinctive regions,
with high sensitivity due to the robots' ability to get close
(within a few cell-diameters) to a source. This capability would
be useful as part of more complex tasks where the robots are to
take some action at such identified regions. Even without
additional actions, the identification itself provides extremely
accurate and rapid diagnostic capability compared to current
technology.

Microscopic robots acting independently to detect specific
patterns of chemicals are analogous to swarms~\cite{bonabeau99}
used in foraging, surveillance or search tasks. Even without
locomotion capabilities, large numbers of such devices could move
through tissues by flowing passively in moving fluids, e.g.,
through blood vessels or with lymph fluid. As the robots move,
they can monitor for preprogrammed patterns of chemical
concentrations.

Chemicals with high concentrations are readily detected with the
simple procedure of analyzing a blood sample. Thus the chemicals
of interest for microscopic robot applications will generally have
low concentrations. With sufficiently low concentrations and small
sources, the devices are likely to only encounter a few molecules
while passing near the source, leading to significant statistical
fluctuations in number of detections.

We can consider this task from the perspective of stages of
operation discussed in the introduction. For a diagnostic first
stage, the robots need only store events in their memory for later
retrieval, when a much more capable conventional computer can
process the information. For an active second stage where robots
react to their detections, e.g., to aggregate at a source location
or release a drug, the robots would need to determine themselves
when they are near a suitable location. In this later case, the
robots would need simple control decision procedure, within the
limits of local information available to them and their
computational capacity. Such a control program could involve
comparing counts to various thresholds, which were determined by
analysis of a previous diagnostic stage of operation.

\subsection{Example Task Environment}

Tissue microenvironments vary considerably in their physical and
chemical properties. As a specific example illustrating the
capabilities of passive motion by microscopic sensors, we consider
a task environment consisting of a macroscopic tissue volume
$\Vtissue$ containing a single microscopic source producing a
particular chemical (or combination of chemicals) while the rest
of the tissue has this chemical at much lower concentrations. This
tissue volume contains a large number of blood vessels, and we
focus on chemical detection in the small vessels, since they allow
exchange of chemicals with surrounding tissue, and account for
most of the surface area. A rough model of the small vessels is
each has length $L$ and they occur throughout the tissue volume
with number density $\rhoVessel$. Localization to volume
$\Vtissue$ could be due to a distinctive chemical environment
(e.g., high oxygen concentrations in the lungs), an externally
supplied signal (e.g., ultrasound) detectable by sensors passing
through vessels within the volume, or a combination of both
methods. The devices are active only when they detect they are in
the specified region.

Robots moving with fluid in the vessels will, for the most part,
be in vessels containing only the background concentration,
providing numerous opportunities for incorrectly interpreting
background concentration as source signals. These false positives
are spurious detections due to statistical fluctuations from the
background concentration of the chemical. Although such detections
can be rare for individual devices, when applied to tasks
involving small sources in a large tissue volume, the number of
opportunities for false positive responses can be orders of
magnitude larger than the opportunities for true positive
detections. Thus even a low false positive rate can lead to many
more false positive detections than true positives. The task
scenario examined in this paper thus includes estimating both true
and false positive rates, addressing the question of whether
simple controls can achieve a good trade-off of both a high true
positive rate and low false positive rate.

For simplicity, we consider a vessel containing only flowing
fluid, robots and a diffusing chemical arising from a source area
on the vessel wall. This scenario produces a static concentration
of the chemical throughout the vessel, thereby simplifying the
analysis. We examine the rate at which robots find the source and
the false positive rate as functions of the detection threshold
used in a simple control rule computed locally by each robot.

\begin{figure}[t]
\begin{center}
\includegraphics[width=\figwidth]{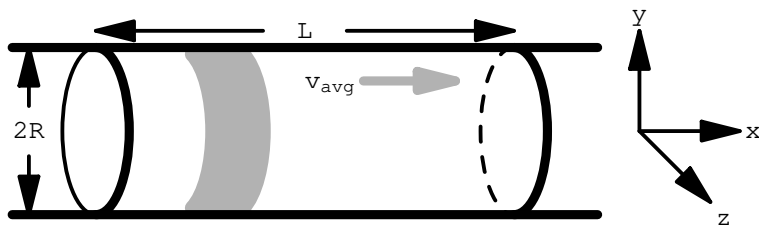}
\end{center}
\caption{\figlabel{schematic}Schematic illustration of the task
geometry as a vessel, of length $L$ and radius $R$. Fluid flows in
the positive $x$-direction with average velocity $\vAvg$. The gray
area is the source region wrapped around the surface of the pipe.}
\end{figure}

\begin{table}[t]
\begin{center}
\begin{tabular}{l|c}
parameter    &   value \\ \hline
\hline \multicolumn{2}{l}{\bf tissue, vessels and source} \\
vessel radius    &   $R=5\micron$   \\
vessel length  &   $L = 1000\micron$    \\
number density of vessels in tissue & $\rhoVessel = 500/\millimeter^3$ \\
tissue volume & $\Vtissue = 1\mbox{cm}^3$ \\
source length   &   $\Ltarget = 30\micron$ \\
\hline \multicolumn{2}{l}{\bf fluid} \\
fluid density   & $\rho=1\mbox{g}/\mbox{cm}^3$ \\
fluid viscosity & $\eta=10^{-2} \mbox{g}/\mbox{cm-s}$ \\
average fluid velocity  &   $\vAvg=1000\micron/\second$ \\
fluid temperature    &   $T=310\mbox{K}$   \\
\hline \multicolumn{2}{l}{\bf robots} \\
robot radius    &   $a=1\micron$   \\
number density of robots   & $\rhoRobot = 200\,\robot/\millimeter^3$ \\
robot diffusion coefficient & $\DRobot = 0.076 \,\micron^2/\second$ \\
\hline \multicolumn{2}{l}{\bf chemical signal} \\
production flux at target  & $\Ftarget = 56 \, \molecule/\second/\micron^2$ \\
diffusion coefficient & $D=100\micron^2/\second$ \\
concentration near source   &  $\Csource = 1.8\, \molecule/\micron^3$ \\
background concentration    &  $\Cbackground = 6\times 10^{-3}\, \molecule/\micron^3$ \\
\end{tabular}
\end{center}
\caption{\tbllabel{parameters}Parameters for the environment,
robots and the chemical signal. The robots are spheres with radius
$a$. The chemical signal concentrations represent a typical 10
kilodalton chemokine molecule, with mass concentrations near the
source and background (i.e., far from the source) equal to
$3\times 10^{-8}\mbox{g/ml}$ and $10^{-10}\mbox{g/ml}$,
respectively. The $\rhoVessel \Vtissue$ small vessels in the
tissue volume occupy about a fraction $\rhoVessel \pi R^2 L
\approx 4\%$ of the volume.}
\end{table}

\fig{schematic} shows the task geometry: a segment of the vessel
with a source region on the wall emitting a chemical into the
fluid. Robots continually enter one end of the vessel with the
fluid flow. We suppose the robots have neutral buoyancy and move
passively in the fluid, with speed given by \eq{fluid velocity} at
their centers. This approximation neglects the change in fluid
flow due to the robots, and is reasonable for estimating detection
performance when the robots are at low enough density to be spaced
apart many times their size, as is the case for the example
presented here. The robot density in \tbl{parameters} corresponds
to $10^{9}$ robots in the entire $5$-liter blood volume of a
typical adult, an example of medical applications using a huge
number of microscopic robots~\cite{freitas99}. These robots use
only about $10^{-6}$ of the vessel volume, far less than the $20\%
- 40\%$ occupied by blood cells. The total mass of all the robots
is about $4\mbox{mg}$.

The scenario for microscopic robots examined here is detecting
small areas of infection or injury. The chemicals arise from the
initial immunological response at the injured area and enter
nearby small blood vessels to recruit white blood
cells~\cite{janeway01}. We consider a typical protein produced in
response to injury, with concentration near the injured tissue of
about $30\mbox{ng}/\mbox{ml}$ and background concentration in the
bloodstream about 300 times smaller. These chemicals, called
chemokines, are proteins with molecular weight around
$10^4\,\mbox{daltons}$. These values lead to the parameters for
the chemical given in \tbl{parameters}, with chemical
concentrations well above the demonstrated sensitivity of
nanoscale chemical sensors~\cite{patolsky05,sheehan05}. This
example incorporates features relevant for medical applications: a
chemical indicating an area of interest, diffusion into flowing
fluid, and a prior background level of the chemical limiting
sensor discrimination.

\subsection{Diffusion of Robots and Chemicals}

Diffusion arising from Brownian motion is somewhat noticeable for
microscopic robots, and significant for molecules. The diffusion
coefficient $D$, depending on an object's size, characterizes the
resulting random motion, with root-mean-square displacement of
$\sqrt{6 D t}$ in a time $t$. For the parameters of
\tbl{parameters}, this displacement for the robots is about $0.7
\sqrt{t}$ microns with $t$ measured in seconds. Brownian motion
also randomly alters robot orientation.

The chemical concentration $C$ is governed by the diffusion
equation~\cite{berg93}
\begin{equation}\eqlabel{diffusion}
\frac{\partial C}{\partial t} = -\nabla \cdot \Flux
\end{equation}
where $\Flux = -D \nabla C + \vFluid C$ is the chemical flux,
i.e., the rate at which molecules pass through a unit area, and
$\vFluid$ is the fluid velocity vector. The first term in the flux
is diffusion, which acts to reduce concentration gradients, and
the second term is motion of the chemical with the fluid.

\begin{figure}[t]
\begin{center}
\includegraphics[width=\figwidth]{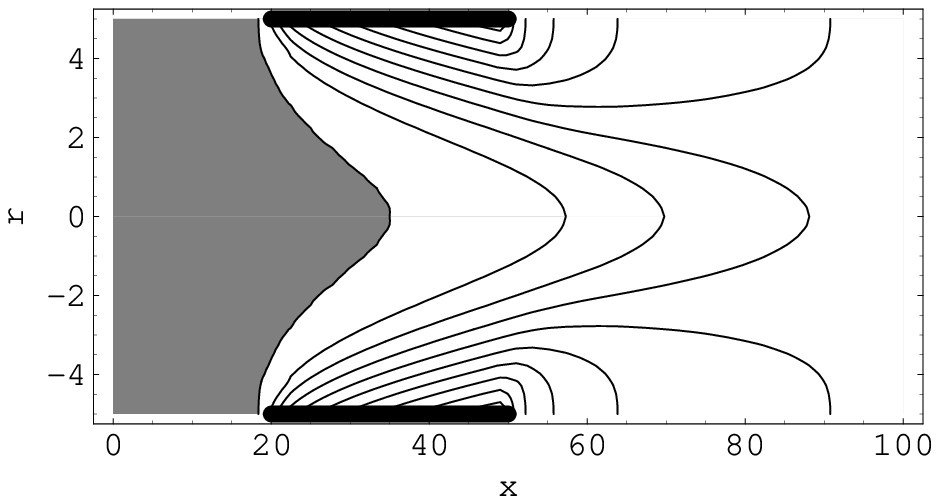}
\end{center}
\caption{\figlabel{concentration}Concentration contours on a cross
section through the vessel, including the axis ($r=0$) in the
middle and the walls ($r=\pm R$) at the top and bottom. The gray
area shows the region where the concentration from the target is
below that of the background concentration. The thick black lines
along the vessel wall mark the extent of the target region. The
vertical and horizontal scales are different: the cross section is
$10\micron$ vertically and $100\micron$ horizontally. Numbers
along the axes denote distances in microns.}
\end{figure}

We suppose the source produces the chemical uniformly with flux
$\Ftarget$. To evaluate high-resolution sensing capabilities, we
suppose the chemical source is small, with characteristic size
$\Ltarget$ as small as individual cells. Total target surface area
is $2 \pi R \Ltarget \approx 940 \micron^2$, about the same as the
surface area of a single endothelial cell lining a blood vessel.
The value for $\Ftarget$ in \tbl{parameters} corresponds to
$5\times 10^4\,\molecule/\second$ from the source area as a whole.
This flux was chosen to make the concentration at the source area
equal to that given in \tbl{parameters}. The background
concentration is the level of the chemical when there is no
injury.

Objects, such as robots, moving in the fluid alter the fluid's
velocity to some extent. For simplicity, and due to the relatively
small volume of the vessel occupied by the robots, we ignore these
changes and treat the fluid flow as constant in time and given by
\eq{fluid velocity}. Similarly, we also treat detection as due to
absorbing spheres with concentration $C$ at the location of the
center of the sphere for \eq{diffusive capture}, assuming the
robot does not significantly alter the average concentration
around it. \fig{concentration} shows the resulting steady-state
concentration from solving \eq{diffusion} with $\partial
C/\partial t = 0$ The concentration decreases with distance from
the source and the high concentration contours occur downstream of
the source due to the fluid flow.

\subsection{Control}

\begin{table}[t]
\begin{center}
\begin{tabular}{l|c}
parameter    &   value \\ \hline
measurement time    &   $\Tmeasure=10\millisecond$   \\
detection threshold & $\Cthreshold$ \\
\end{tabular}
\end{center}
\caption{\tbllabel{control parameters}Robot control parameters for
chemical signal detection.}
\end{table}

The limited capabilities of the robots and the need to react on
millisecond time scales leads to emphasizing simple controls based
on local information. For chemical detection at low
concentrations, the main sensor limitation is the time required
for molecules to diffuse to the sensor. Thus the detections are a
stochastic process. A simple decision criterion for a robot to
determine whether it is near a source is if a sufficient number of
detections occur in a short time interval. Specifically, a robot
considers a signal detected if the robot observes at least
$\Cthreshold$ counts in a measurement time interval $\Tmeasure$.
The choice of measurement time must balance having enough time to
receive adequate counts, thereby reduce errors due to statistical
fluctuations, while still responding before the robot has moved
far downstream of the source where response would give poor
localization of the source or, if robots are to take some action
near the source, require moving upstream against the fluid flow.
Moreover, far downstream of the source the concentration from the
target is small so additional measurement time is less useful. A
device passing through a vessel with a source will have about
$\Ltarget/\vAvg = 30 \millisecond$ with high concentration, so a
measurement time of roughly this magnitude is a reasonable choice,
as selected in \tbl{control parameters}. A low value for
$\Cthreshold$ will produce many false positives, while a high
value means many robots will pass the source without detecting it
(i.e., false negatives). False negatives increase the time
required for detection while false positives could lead to
inappropriate subsequent activities (e.g., releasing a drug to
treat the injury or infection at a location where it will not be
effective).

\subsection{Analysis of Behavior}

Task performance depends on the rate robots detect the source as
they move past it, and the rate robots incorrectly conclude they
detect the source due to background concentration of the chemical.

From the values of \tbl{parameters}, robots enter any given vessel
at an average rate
\begin{equation}\eqlabel{sensorRateOne}
\sensorRateOne = \rhoRobot \pi R^2 \vAvg \approx 0.016/\second
\end{equation}
and the rate robots enter (and leave) any of the small vessels
within the tissue volume is
\begin{equation}\eqlabel{sensorRate}
\sensorRate = \rhoVessel \Vtissue \sensorRateOne \approx 8\times
10^3/\second
\end{equation}
A robot encounters changing chemical concentration as it moves
past the source. The expected number of counts a robot at position
$\position$ has received from the source chemical during a prior
measurement interval time $\Tmeasure$ is
\begin{equation}\eqlabel{K}
K(\position) = 4 \pi D a \int_0^{\Tmeasure}
C(\position'(\position,\tau)) d \tau
\end{equation}
where $\position'(\position,\tau)$ denotes the location the robot
had at time $\tau$ in the past. During the time the robot passes
the target, Brownian motion displacement is $\sim 0.1\micron$,
which is small compared to the $10\micron$ vessel diameter. Thus
the possible past locations leading to $\position$ are closely
clustered and for estimating the number of molecules detected
while passing the target, a reasonable approximation is the robot
moves deterministically with the fluid. In our axially symmetric
geometry with fluid speed given by \eq{fluid velocity}, positions
are specified by two coordinates $\position=(r,x)$ so
$\position'((r,x),\tau) = (r,x-v(r) \tau)$ when the robot moves
passively with the fluid and Brownian motion is ignored. During
this motion, the robot will, on average, also encounter
\begin{equation}\eqlabel{k}
k = 4 \pi D a c \Tmeasure
\end{equation}
molecules from the background concentration, not associated with
the source.

With diffusive motion of the molecules, the actual number of
counts is a Poisson distributed random process. The detection
probability, i.e., having at least $E$ events when the expected
number is $\mu$, is
\begin{displaymath}
\Pr(\mu,E) = 1 - \sum_{n=0}^{E-1} \Poisson(\mu,n)
\end{displaymath}
where $\Poisson(\mu,n)=e^{-\mu}\mu^n/n!$ is the Poisson
distribution.

Taking the devices to be ideal absorbing spheres for the chemical
described in \tbl{parameters}, \eq{diffusive capture} gives the
capture rates $\gamma \approx 8/\second$ at the background
concentration and $\approx 2300/\second$ near the source.
Detection over a time interval $\Delta t$ is a Poisson process
with mean number of detections $\mu = \gamma \Delta t$. Consider a
robot at $\position$. During a small time interval $\Delta t$ the
probability to detect a molecule is $4 \pi D a (C(\position)+c)
\Delta t \ll 1$. For a robot to first conclude it has detected a
signal during this short time it must have $\Cthreshold-1$ counts
in the prior $\Tmeasure-\Delta t$ time interval and then one
additional count during $\Delta t$. Thus the rate at which robots
first conclude they detect a signal is
\begin{equation}\eqlabel{detection rate}
4 \pi D a (C+c)
\frac{\Poisson(K+k,\Cthreshold-1)}{\sum_{n<\Cthreshold}\Poisson(K+k,n)}
\end{equation}
In \eq{detection rate} $C$ and $K$ depend on robot position and
the last factor is the probability the robot has $\Cthreshold-1$
counts in its measurement time interval, given it has not already
detected the signal, i.e., the number of counts is less than
$\Cthreshold$. This expression is an approximation: ignoring
correlations in the likelihood of detection over short time
intervals. \eq{detection rate} also gives the detection rate when
there is no source, i.e., false positives, by setting $C$ and $K$
to zero.

To evaluate the rate robots detect the source as they pass it, we
view the robots as having two internal control states: MONITOR and
DETECT. Robots are initially in the MONITOR state, and switch to
the DETECT state if they detect the chemical, i.e., have at least
$\Cthreshold$ counts during time $\Tmeasure$. Using the stochastic
analysis approach to evaluating robot behavior, the steady-state
concentrations of robots monitoring for the chemical, $\Rmonitor$,
is governed by \eq{diffusion} for $\Rmonitor$ instead of chemical
concentration, with the addition of a decay due to robots changing
to the DETECT state, i.e.,
\begin{equation}\eqlabel{signal}
\nabla \cdot (\DRobot \nabla \Rmonitor - \vFluid \Rmonitor) -
\aMonitorToDetected \Rmonitor = 0
\end{equation}
The signal detection transition, $\aMonitorToDetected$, is given
by \eq{detection rate} and depends on the choice of threshold
$\Cthreshold$ and robot position.

The rate sensors detect the source using a threshold $\Cthreshold$
is
\begin{equation}
\sourceRate = \sensorRateOne \int \aMonitorToDetected \Rmonitor dV
\end{equation}
where $\sensorRateOne$ is given by \eq{sensorRateOne} and the
integral is over the interior volume of the vessel containing the
source.

The background concentration can give false positives, i.e.,
occasionally producing enough counts to reach the count threshold
in time $\Tmeasure$. The background concentration extends
throughout the tissue volume giving many opportunities for false
positives. With the parameters of \tbl{parameters}, the expected
count from background in $\Tmeasure$ is $\Ebackground = 0.08$.
Since a sensor spends $\approx L/\vAvg = 1\second$ in a small
vessel in the tissue volume, the sensor has about 100 independent
$10\millisecond$ opportunities to accumulate counts toward the
detection threshold $\Cthreshold$. The rate of false positive
detections is then
\begin{equation}
\backgroundRate \approx 100\; \sensorRate
\Pr(\Ebackground,\Cthreshold)
\end{equation}

For a diagnostic task, we can pick a detection threshold
$\Cthreshold$ and a time $\Ttask$ for sensors to accumulate
counts. The expected number of sensors reporting detections from
the source and from the background are then $\sourceRate \Ttask$
and $\backgroundRate \Ttask$, respectively. The actual number is
also a Poisson process, so another decision criterion for
declaring a source detected is the minimum number of sensors $n$
reporting a detection. Since expected count rate near the source
is significantly larger than the background rate, the
contributions to the counts from the source and background are
nearly independent, so the probability for detecting a source is
\begin{displaymath}
\Pr\left((\sourceRate+\backgroundRate) \Ttask,n\right)
\end{displaymath}
and similarly for the false positives with counts based only on
$\backgroundRate$.

\subsection{Detection Performance}

\begin{figure}
\begin{center}
\includegraphics{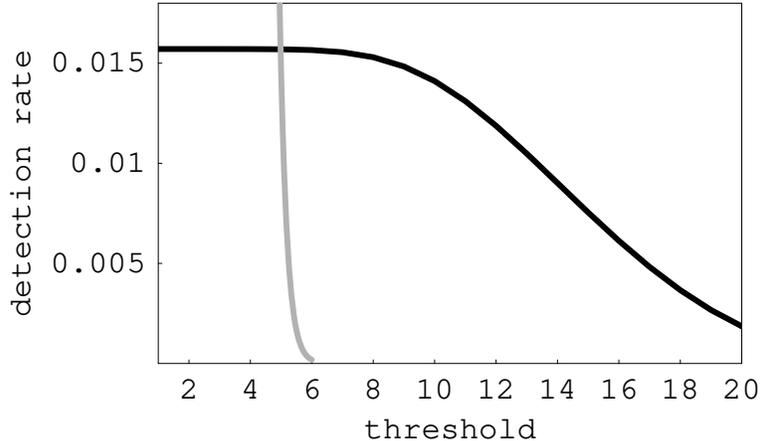}
\end{center}
\caption{Rates robots detect a signal for those passing through
the single small vessel with the source (black) and all the others
in the tissue volume, i.e., the false positives (gray) as a
function of threshold $\Cthreshold$ used for detection during the
$\Tmeasure$ time interval.}\figlabel{detection rates}
\end{figure}

\fig{detection rates} shows the values of $\sourceRate$ and
$\backgroundRate$ for the parameters of \tbl{parameters} for
various choices of the control parameter $\Cthreshold$. Despite
the much larger number of opportunities for false positives
compared to the single vessel with the source, the ability of
robots to pass close to the source allows selecting $\Cthreshold
\approx 10$ for which false positive detections are small while
still having a significant rate of true positives.

\begin{figure}
\begin{center}
\includegraphics{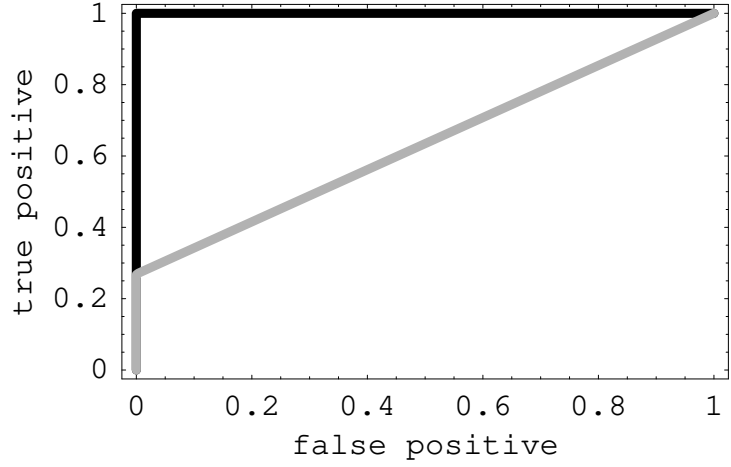}
\end{center}
\caption{Probabilities of detecting a single source (true
positive) and mistaking background concentration for a source
(false positive) after sampling for $\Ttask=20$ and 1000 seconds
(gray and solid curves, respectively). Each curve corresponds to
changing the minimum threshold $\Cthreshold$ of counts during the
time interval $\Tmeasure$ required to indicate a detection
event.}\figlabel{detection}
\end{figure}

For diagnostics, an indication of performance is comparing the
likelihood of true and false positives. In particular, identifying
choices of control parameters giving both a high chance of
detecting a source and a low chance of false positives.
\fig{detection} illustrates the tradeoff for the task considered
here. The curves range from the lower-left corner (low detection
rates) with a high threshold to the upper-right corner (high
detection and high false positive rate) with a low threshold.
Robots collecting data for only about 20 minutes allow high
performance, in this case with $\Cthreshold$ around 10. This
corresponds to the behavior seen in \fig{detection rates}:
$\Cthreshold \approx 10$ is high enough to be rarely reached with
background concentration alone (in spite of the much larger number
of vessels without the source than the single vessel with the
source), but still low enough that most devices passing through
the single vessel with the source will detect it.

\begin{figure}
\begin{center}
\includegraphics{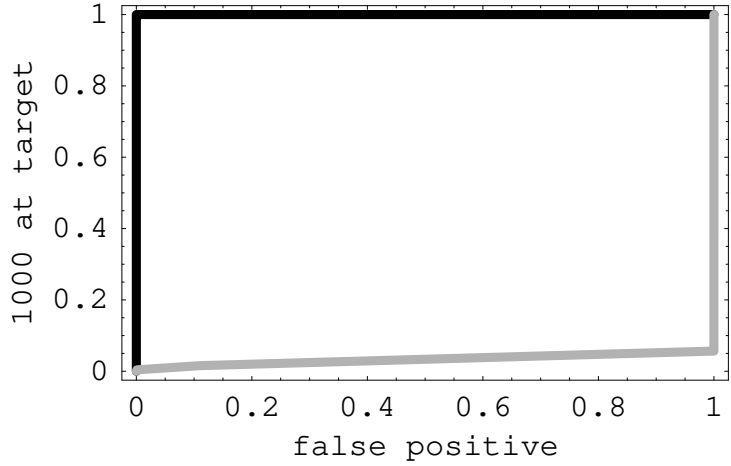}
\end{center}
\caption{Probability of at least 1000 robots detecting the source
vs. the probability at least one robot mistakes background
concentration for a source (false positive) after sampling for
$\Ttask=6\times 10^4$ and $10^5$ seconds (gray and solid curves,
respectively). Each curve corresponds to changing the minimum
threshold $\Cthreshold$ of counts during the time interval
$\Tmeasure$ required to indicate a detection
event.}\figlabel{detection1000}
\end{figure}

If the robots are to act at the source, ensuring at least one
detection may not be enough. For instance, if the robots release a
chemical near the source, or aggregate near the source (e.g., to
stick to vessel wall near the source to provide enhanced imaging
or to mechanically alter the tissue), then it may be necessary to
ensure a relatively large number of robots detect the source. At
the same time, we wish to avoid inappropriate actions due to false
positives. A criterion emphasizing safety is having a high chance
the required number detect the source before a single robot has a
false positive detection, so not even a single inappropriate
action takes place. \fig{detection1000} shows we can achieve high
performance by this criterion: the robots circulate for about a
day to have enough time for 1000 to detect the source while still
having a low chance for any false positives.

Another motivation for requiring more than one detection at the
source is to account for sensor failures, e.g., requiring
detection by several sensors as independent confirmation of the
source. Occasional spurious extra counts by the sensors amount to
an increase in the effective background concentration. As long as
these extra counts are infrequent, and not significantly clustered
in time, such errors will not significantly affect the overall
accuracy of the results.

In summary, simple control allows fast and accurate detection of
even a single cell-sized source within a macroscopic tissue
volume. The key feature enabling this performance is the robots'
ability to pass close to individual cells.

For comparison, instead of using microscopic sensors, one could
attempt to detect the chemical from a blood sample. This allows
using chemical sensors outside the body, giving simpler
fabrication and use. However, such a sample dilutes the chemical
throughout the blood volume, resulting in considerably smaller
concentrations than are available to microscopic sensors passing
close to the source. As an example, suppose a single source
described above produces the chemical for one day and all this
production is delivered to the blood without any degrading before
a sample is taken. The source producing $\sim 5\times 10^4
\molecule/\second$ gives a concentration in the 5 liter blood
volume of about $7 \times 10^{11} \molecule/\meter^3$. This value
is about $10^{-4}$ of the background concentration, so the
additional chemical released by the source would be difficult to
detect against small variations in background concentration.

These performance estimates also indicate behavior in other
scenarios. For instance, with fewer sensors, detection times would
be correspondingly longer, or would only be sensitive to a larger
number of sources. For instance, with $10^6$ sensors, a factor of
1000 fewer than in \tbl{parameters}, achieving the discrimination
shown in \fig{detection} would take a thousand times longer.
Alternatively, for $10^6$ sensors with 1000 sources distributed
randomly in the tissue volume instead of just one source,
performance would be similar to that shown in the figure. The
stochastic analysis approach used here could also estimate other
aspects of robot performance, such as the average distance to the
source if and when a passing robot detects it.

\begin{figure}
\begin{center}
\includegraphics{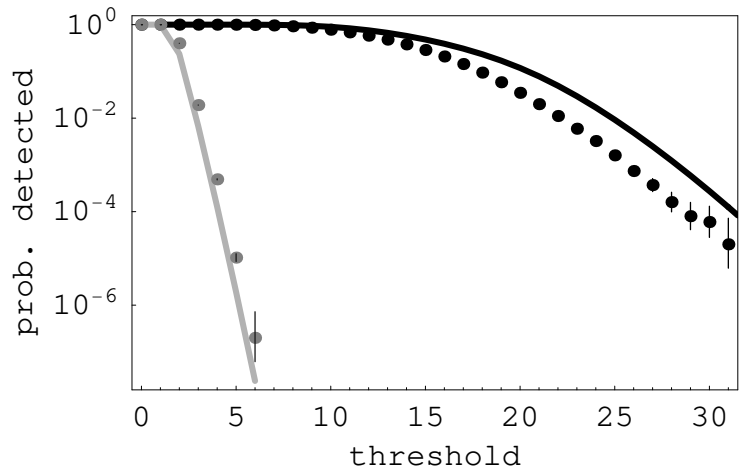}
\end{center}
\caption{Chemical detection probabilities vs. threshold
$\Cthreshold$ for detection during the $\Tmeasure$ time interval.
Two cases are shown: a robot passing the source (black) and a
robot in a small vessel without the source, i.e., false positive
detection (gray). Curves are the theoretical estimates, and points
are from simulations. Each point includes a line showing the 95\%
confidence interval of the probability estimate, which in most
cases is smaller than the size of the plotted point. The points
are averages from $10^5$ and $10^7$ discrete simulations of single
robots passing through vessels with and without a source,
respectively.}\figlabel{detection probability}
\end{figure}

The stochastic analysis approach to evaluating robot behaviors in
spatially varying fields makes various simplifying approximations
to obtain $\sourceRate$ and $\backgroundRate$. These include
independence of the number of detection counts in different time
intervals, characterizing the robots by a continuous concentration
field rather than discrete objects, and ignoring how the robots
(and other objects, such as blood cells) change the fluid flow and
the concentration distribution of the chemical in the fluid. In
principle, the analysis approach can readily include objects in
the fluid, but the numerical solution becomes significantly more
complicated. Instead of the parabolic velocity profile \eq{fluid
velocity}, fluid flow changes with time as the objects move
through the vessel. Simultaneously, the object motion is
determined by the drag force from the fluid. Evaluating the flow
requires solving the Navier-Stokes equation, which can also
include more complicated geometries for the vessels and source
than treated here. Despite this additional numerical complexity,
analyzing robot behaviors is formally the same.

On the other hand, history-dependent behavior of the robots, such
as checking for a certain number of counts within a specified time
interval, is difficult to include in the formalism, leading to the
simplifying independence assumptions used here. As a check on the
accuracy of these assumptions, a discrete-event simulation for the
same task evaluates robot behavior without making those
assumptions, but requires significantly more compute time to
evaluate average behaviors. \fig{detection probability} shows this
comparison for the key quantities used here: the detection
probabilities for true and false positives. We see a fairly close
correspondence between the two methods, but with a systematic
error.

\section{Applications for Additional Robot Capabilities}

High-resolution in vivo chemical sensing with passive motion in
fluids is well-suited to robots with minimal capabilities.
Additional hardware capabilities allow collecting more information
communicating with external observers during operation, or taking
actions such as aggregating at the chemical source, releasing
chemicals or mechanically manipulating the cells. This section
presents a number of such possibilities.

\subsection{Improved Inference from Sensor Data}

The example in \fig{detection} uses a simple detection criterion
based on a single choice of threshold $\Cthreshold$. More
sophisticated criteria could improve accuracy. One example is
matching the temporal distribution of detections to either that
expected from a uniform background or a spatially localized high
concentration source. This procedure would also be useful when the
source and background concentrations are not sufficiently well
known a priori to determine a suitable threshold. Instead, the
distribution of counts could distinguish the low background rate
encountered most of the time from occasional clusters of counts at
substantially higher rates.

Robots could use correlations in space or time for improved
sensitivity. For instance, if a source emits several chemicals
together, each of which has significant background concentration
from a variety of separate sources, then detecting all of them
within a short period of time improves statistical power of the
inference. Similarly if the chemical is released in bursts,
sensors nearby during a burst would encounter much higher local
concentrations than the time averaged concentration. Stochastic
temporal variation in protein production is related to the
regulatory environment of the genes producing the protein, in
particular the number of proteins made from each
transcript~\cite{mcadams97}. Thus temporal information could
identify aspects of the gene regulation within individual cells.

Correlations in the measurements could distinguish between a
strong source and many weak sources throughout the tissue volume
producing the chemical at the same total rate. The strong source
would give high count rates for a few devices (those passing near
the source) while multiple weak sources would have some detection
in a larger fraction of the sensors.

As another example, using fluid flow sensors would allow
correlating chemical detections with properties of the flow and
the vessel geometry (e.g., branching and changes in vessel size or
permeability to fluids).

\subsection{Correlating Measurements from Multiple Devices}

Determining properties of the chemical sources would improve if
data from devices passing different sources is distinguished from
multiple devices passing the same source. One possibility for
correlating information from different devices is if sources, or
their local environments, are sufficiently distinct chemically so
measuring similar ratios of various chemicals in different devices
likely indicates they encountered the same source. Common temporal
variation could also suggest data from the same source.

With less distinctive sources or insufficient time to collect
enough counts to make the distinction, devices capable of
communicating with others nearby could provide this correlation
information directly. With the parameters of \tbl{parameters},
typical spacing between devices is several hundred microns, which
is beyond a plausible communication distance between devices. Thus
direct inter-device communication requires reducing the spacing
with either a larger total number of devices or temporary local
aggregation in small volumes.

Devices could achieve this aggregation if they can alter their
surface properties to stick to the vessel wall~\cite{lahann05}.
With such a programmable change, a device detecting an interesting
event could stick when it next encounters the wall. The parameters
of \tbl{parameters} give, on average, about one device passing
through a given small vessel each minute. Thus a device on the
wall could remain there for a few minutes, broadcasting its unique
identifier to other devices as they pass. Devices would record the
identifiers they receive, with a time stamp, thereby correlating
their detections.

Further inferences could be made from devices temporarily on the
vessel wall where small vessels merge into larger ones. In this
case, the identifier received from a device on the wall enables
correlating events in nearby vessels, i.e., those that merge into
the same larger vessel. Similarly, if robots aggregate in upstream
vessels, before they branch into smaller ones, the robots could
record each others' unique identifiers. Subsequent measurements
over the next few hundred milliseconds would be known to arise
from the same region, i.e., either the same vessel or nearby ones.

The ability to selectively stick to vessel walls would allow
another mode of operation: the devices could be injected in larger
blood vessels leading into a macroscopic tissue volume of interest
and then stick to the vessel walls after various intervals of time
or when specific chemicals are detected. After collecting data,
the devices would release from the wall for later retrieval.

\subsection{Reporting During Operation}

The devices could carry nanoscale structures with high response to
external signals. Such structures could respond to light of
particular wavelengths when near the skin~\cite{wang05a}, or give
enhanced imaging via MRI or ultrasound~\cite{liu06}. Such
visualization mechanisms combined with a selective ability to
stick to vessel walls allows detecting aggregations of devices at
specified locations near the surface of the body~\cite{service05}.

This visualization technique could be useful even if the tissue
volume of interest is too deep to image effectively at high
resolution. In particular, robots could use various areas near the
skin (e.g., marked with various light or ultrasound frequencies)
at centimeter scales as readout regions during operation. Devices
that have detected the chemicals could aggregate at the
corresponding readout location, which would then be visible
externally. Devices could choose how long to remain at the
aggregation points based on how high a concentration of the
chemical pattern they detected. This indication of whether, and
(at a coarse level) what, the devices have found could help decide
how long to continue circulating to improve statistics for weak
chemical signatures. These aggregation points could also be used
to signal to the devices, e.g., instructing them to select among a
few modes of operation.

\subsection{Detection of Chemicals inside Cells}

The task described above relies on detecting chemicals released
into the bloodstream. However, some chemicals of interest may
remain inside cells, or if released, be unable to get into the
bloodstream. In that case, sensors in the bloodstream would not
detect the chemical even when they pass through vessels near the
cells.

However, an extension of the protocol could allow indirect
detection of intracellular chemicals. For example, current
technology can create molecules capable of entering cells, and, if
they encounter specific chemicals, changing properties to emit
signals or greatly enhance response to external imaging methods.
Such molecules can indicate a variety of chemical behaviors within
cells~\cite{sunney06}. Thus if the microscopic devices include
sensors for these indicator signals, they could indirectly record
the activity of the corresponding intracellular chemicals in
nearby cells. Such sensing from within nearby blood vessels would
complement current uses of these marker molecules with much larger
scale whole body imaging. In this extended protocol, the
microscopic devices would provide the same benefits of detecting
chemicals directly by instead detecting a proxy signal that is
able to reach the nearby blood vessels.

\subsection{Modifying Microenvironments}

Beyond the diagnostic task discussed above, robots able to locate
chemically distinctive microenvironments in the body could have
capabilities to modify those environments. For instance, the
devices could carry specific drugs to deliver only to cells
matching a prespecified chemical
profile~\cite{freitas99,freitas06} as an extension of a recent in
vitro demonstration of this capability using DNA
computers~\cite{benenson04}.

With active locomotion, after detecting the chemicals the devices
could follow the chemical gradient to the source, though this
would require considerable energy to move upstream against the
fluid flow. Alternatively, with sufficient number of robots so
they are close enough to communicate, a robot detecting the
chemical could acoustically signal upstream devices to move toward
the vessel wall. In this cooperative approach, the detecting
device does not itself attempt to move to the source, but rather
acts to signal others upstream from the source to search for it.
These upstream devices would require little or no upstream motion.
Furthermore, with a large number of devices, even if only a small
fraction move in the correct direction to the source after
receiving a signal, many would still reach the source. This
approach of using large numbers and randomness in simple local
control is analogous to that proposed for collections of larger
reconfigurable robots~\cite{bojinov02,rus99,salemi01}. The
behavior of microscopic active swimmers~\cite{dreyfus05} raises
additional control issues to exploit the hydrodynamic interactions
among swimming objects as they aggregate so the distance between
devices becomes only a few times their size~\cite{hernandez05}.

Robots aggregated at chemically identified targets could perform
precise microsurgery at the scale of individual cells. Since
biological processes often involve activities at molecular, cell,
tissue and organ levels, such microsurgery could complement
conventional surgery at larger scales. For instance, a few
millimeter-scale manipulators, built from micromachine (MEMS)
technology, and a population of microscopic devices could act
simultaneously at tissue and cellular size scales. An example
involving microsurgery for nerve repair with plausible biophysical
parameters indicates the potential for significant improvement in
both speed and accuracy compared to the larger-scale machines
acting alone~\cite{sretavan05,hogg05}.

\section{Discussion}

Plausible capabilities for microscopic robots suggest a range of
novel applications in biomedical research and medicine. Sensing
and acting with micron spatial resolution and millisecond timing
allows access to activities of individual cells. The large numbers
of robots enable such activities simultaneously on a large
population of cells in multicellular organisms. In particular, the
small size of these robots allows access to tissue through blood
vessels. Thus a device passes within a few tens of microns of
essentially every cell in the tissue in a time ranging from tens
of seconds to minutes, depending on the number of devices used.

With many devices in the tissue but only a few in the proper
context to perform task, false positives are a significant issue.
In some situations, these false positives may just amount to a
waste of resources (e.g., power). But in other cases, too many
false positives could be more serious, e.g., leading to
aggregation blocking blood vessels or incorrect diagnosis.

The sensing task described in this paper highlights key control
principles for microscopic robots. Specifically, by considering
the overall task in a series of stages, the person using the
devices remains in the decision loop, especially for the key
decision of whether to proceed with manipulation (e.g., release a
drug) based on diagnostic information reported by the devices.
Information retrieved during treatment can also indicate whether
the procedure is proceeding as planned and provide high-resolution
documentation.

The performance estimates for the sensing task show devices with
limited capabilities -- specifically, without locomotion or
communication with other devices -- can nevertheless rapidly
detect chemical sources as small as a single cell. The devices use
their small size and large numbers to allow at least a few to get
close to the source, where concentration is much higher than
background. This paper also illustrates use of an analysis
technique for average behavior of microscopic robots that readily
incorporates spatially variable fields in the environment. Such
fields are of major significance for microscopic robots, in
contrast to their usual limited importance for larger robots.

As a caveat on the the results, the model examined here treats the
location of the chemical source as independent of the properties
and flow rates in the vessel containing the source. Systematic
variation in the density and organization of the vessels will
increase the variation in detected values. For instance, the
tissue could have correlations between vessel density and the
chemical sources (e.g., if those chemicals enhance or inhibit
growth of new vessels). Accurate inference require models of how
the chemicals move through the tissue to nearby blood vessels.
Chemicals could react after release from the source to change the
concentrations with distance from the source. Nevertheless, the
simple model discussed here indicates the devices could have high
discrimination for sources as small as single cells. Thus even
with some unmodeled sources of variability, good performance could
still be achieved by extending the sensing time or using more
sophisticated inference methods. Moreover, with some localization
during operation, the devices themselves could estimate some of
this variation (e.g., changes in density of vessels in different
tissue regions), and these estimates could improve the inference
instead of relying on average or estimated values for the tissue
structure.

Further open questions include the effect of higher diffusion from
mixing due to motion of cells in fluid, for both chemicals and
robots. For instance, the hydrodynamic effect of blood cells
moving in the fluid greatly increases the diffusion coefficient of
smaller objects in the fluid, to about $1000
\micron^2/\second$~\cite{keller71}.

Instead of flowing with fluid, the sensors could be implanted at
specific locations of interest to collect data in their local
environments, and later retrieved. This approach does not take
full advantage of the sensor size: it could be difficult to
identify interesting locations at cell-size resolutions and
implant the devices accurately. Nevertheless, such implants could
be useful by providing local signals to indicate regions of
interest to other sensors passing nearby in a moving fluid.

Safety is important for medical applications of microscopic
robots. Thus, evaluating a control protocol should consider its
accuracy allowing for errors, failures of individual devices or
variations in environmental parameters. For the simple distributed
sensing discussed in this paper, statistical aggregation of many
devices' measurements provides robustness against these
variations, a technique recently illustrated using DNA computing
to respond to patterns of chemicals~\cite{benenson04}.
Furthermore, the devices must be compatible with their biological
environment~\cite{nel06} for enough time to complete their task.
Appropriately engineered surface coatings and
structures~\cite{freitas03} should prevent unwanted inflammation
or immune system reactions during robot operation. However, even
if individual devices are inert, too many in the circulation would
be harmful. From \tbl{parameters}, sensors occupy a fraction
$(4/3)\pi a^3 \rhoRobot \approx 10^{-6}$ of the volume inside the
vessels. This value is well below the fraction, about $10^{-3}$,
of micron-size particles experimentally demonstrated to be safely
tolerated in the circulatory system of at least some
mammals~\cite{freitas03}. Thus the number of sensors used in the
protocol of this paper is unlikely to be a safety issue.

Despite the simplifications used to model sensor behavior, the
estimates obtained in this paper with plausible biophysical
parameters show high-resolution sensing is possible with passive
device motion in the circulatory system, even without
communication capabilities. Thus relatively modest hardware
capabilities could provide useful in vivo sensing capabilities.
Research studies of tissue microenvironments with such devices
will enable better inferences from their data and indicate
distributed controls suitable for more capable devices.

\section*{Acknowledgments}
I have benefited from discussions with Philip J. Kuekes and David
Sretavan.

\end{document}